\documentclass{article} 
\usepackage[final]{colm2026_conference}

\usepackage{microtype}
\usepackage{hyperref}
\usepackage{url}
\usepackage{booktabs}
\usepackage{graphicx}
\usepackage{subcaption}
\usepackage{wrapfig}

\usepackage{amsmath,amsfonts,bm}

\def\eqref#1{equation~\ref{#1}}

\def\1{\bm{1}}

\DeclareMathAlphabet{\mathsfit}{\encodingdefault}{\sfdefault}{m}{sl}
\SetMathAlphabet{\mathsfit}{bold}{\encodingdefault}{\sfdefault}{bx}{n}

\usepackage{lineno}

\definecolor{darkblue}{rgb}{0, 0, 0.5}
\hypersetup{colorlinks=true, citecolor=darkblue, linkcolor=darkblue, urlcolor=darkblue}

\title{Evaluating Steering Techniques using Human Similarity \\Judgments}

\author{
Zach Studdiford$^{1}$ \quad Timothy T. Rogers$^{1}$ \quad Siddharth Suresh$^{*1}$ \quad Kushin Mukherjee$^{*2}$ \\[0.5ex]
\textnormal{$^{1}$University of Wisconsin--Madison \quad $^{2}$Stanford University} \\[0.5ex]
\texttt{\{studdiford, ttrogers, siddharth.suresh\}@wisc.edu} \\
\texttt{kushinm@cs.stanford.edu}
}

\begin{document}

\ifcolmsubmission
\linenumbers
\fi

\maketitle
\newcommand{\zs}[1]{\textcolor{red}{ZS: #1}}

\begin{abstract}

Current evaluations of Large Language Model (LLM) steering techniques focus on task-specific performance, overlooking how well steered representations align with human cognition.
Using a well-established triadic similarity judgment task, we assessed steered LLMs on their ability to flexibly judge similarity between concepts based on size or kind, two central dimensions organizing human mental representations. We found that prompt-based steering methods outperformed other methods both in terms of steering accuracy and model-to-human alignment. We also found LLMs were biased towards `kind' similarity and struggled with `size' alignment. This evaluation approach, grounded in human cognition, adds further support to the efficacy of prompt-based steering and reveals privileged representational axes in LLMs prior to steering.

\end{abstract}
\section{Introduction}

The ability to marshal one's knowledge of the world, while taking into account the present context, in service of guiding behavior is what makes human cognition remarkably flexible
\citep{botvinick2014computational, giallanza2024integrated, wong2025modeling}. Different aspects of a concept can be preferentially activated to execute different tasks. For example, the \textit{size} of an orange might be important when a shopper is deciding how many can fit inside their shopping basket, whereas the \textit{kind} of produce it is (i.e., fruit) might be important when a grocer is arranging their wares on a shelf. Such flexible behavior is facilitated by learning robust representations of concepts, captured by theories and models of semantic knowledge \citep{rogers2004semantic, rogers2024generalization, saxe2019mathematical}, and by `guiding' these representations through context-sensitive mechanisms, captured by accounts of cognitive/semantic control \citep{cohen1990control, miller2001integrative, lambonralph2017neural, giallanza2024integrated}.

These aspects of human semantic cognition---representation and control---provide strong analogies to two important axes of evaluation for large language models (LLMs): (1) representation learning through pre-training and (2) intervention-based steering. This isomorphism is important insofar as cognitive scientists have developed methods for characterizing human mental representations under varied contexts for naturalistic tasks that underpin a variety of human behaviors.
Thus, these same techniques can be brought to bear to shed light on how representation and control manifest in LLMs. Here, we focus on evaluating current LLM steering techniques, i.e., techniques for controlling LLM representations, through the lens of cognitive science-inspired methods.
This approach constitutes a critical complement to efforts in interpretability research, primarily stemming from the computer sciences. A key factor in our approach is that it allows for evaluating steering techniques not just in terms of performance, but in terms of how aligned the resultant model behaviors and representations are to those of humans when presented with qualitatively similar interventions. In other words, we ask - ``Do steering-based methods transform models' internal representations the same way in which context transforms humans' mental representations?''.

In the present work, we used a triadic similarity judgment task \citep{sievert2023efficiently, hebart2020revealing}, a technique that has been shown to be effective at characterizing human mental representations, where humans judged the similarity between concepts in terms of either \textbf{size} or \textbf{kind}. We then tested a suite of LLM intervention steering techniques on \texttt{gemma-2-27b-it}, \texttt{gemma-2-9b-it}, and \texttt{qwen-2.5-14b} and assessed each method's (1) \emph{accuracy}, measured as the number of `correct' judgments on the size and kind tasks, and (2) \emph{human alignment}, measured as the Procrustes correlation between embeddings generated from human judgments versus those generated from model judgments. Model accuracy differed dramatically across steering methods, with some approaches yielding human-level performance; but surprisingly, human alignment was poor, especially for size judgments across all steering methods. These results suggest critical differences between semantic control in humans and large language models and opportunities to bring these two systems into closer alignment.

\section{Related work}

\subsection{Triadic judgment tasks} 
Triadic judgment tasks, coupled with advances in embedding algorithms have enabled the characterization of the representational geometry underlying human concepts \citep{jamieson2015next, sievert2023efficiently, muttenthaler2022vice, hebart2020revealing, muttenthaler2023improving, suresh2024categories, giallanza2024integrated}. The general approach is to present participants with three concepts and to ask them to indicate either the odd one out or which of two options is most similar to a designated target. The judgments are used to estimate a low-dimensional embedding where similar concepts are closer together. Such techniques have been extended to AI systems, specifically vision models \citep{muttenthaler2023human, mukherjee2025using}, to estimate how human-like neural network representations are (see \citeauthor{sucholutsky2023getting}, \citeyear{sucholutsky2023getting}). Judgments made by LLMs on such tasks can also be used to estimate model embeddings, helping uncover representational structures in otherwise opaque systems \citep{hebart2020revealing, sucholutsky2023getting}. This approach has revealed fundamental differences: human conceptual structures remain consistent across cultures, while LLMs exhibit task-dependent variation \citep{suresh2023conceptual}. Applications extend to standardized similarity norms \citep{jiang2022visual} and semantic organization principles \citep{mirman2017taxonomic}.

\subsection{Model steering methods} 

Although language models are already context-sensitive due to their outputs being conditioned on prior text, steering methods further control behavior via internal interventions or structured prompts. Because these methods allow for fine-grained control over LLM behavior at inference time, they offer a potential alternative to more expensive methods--such as supervised fine tuning (SFT), which is computationally expensive and directly modifies model weights \citep{wu2025axbench}. 

\paragraph{Prompting}
In many instances, model outputs can be steered directly through prompt instructions provided in natural language \citep{sahoo2025systematicsurveypromptengineering}. \textit{Zero-shot prompting} involves providing a single task instruction or example to direct LLM outputs, and is capable of shaping responses to support emergent behaviors including multi-step reasoning \citep{kojima2023largelanguagemodelszeroshot}, social simulation \citep{chuang2024demographicsaligningroleplayingllmbased} and image classification \citep{abdelhamed2025seeenhancingzeroshotimage}. Even without explicit instruction, LLMs can induce general rules and abstractions from minimal examples via \textit{in-context learning}. This is sufficient to steer model behavior for translation tasks, \citep{brown2020languagemodelsfewshotlearners}, logical reasoning \citep{yin2024deeperinsightsupdatespower}, and task induction \citep{honovich2022instructioninductionexamplesnatural}.

\paragraph{Task vectors}

\citet{hendel2023incontext} find that LLMs encode in-context task instructions as a linear direction in intermediate layers. This direction, $\theta$, is a compressed representation of the abstract rule underlying the sequence. When $\theta$ is extracted from the forward pass on a rule-based prompt $x$ and patched into the forward pass of an empty prompt $x'$ with no task instructions, the model is able to produce the correct sequence completion for the empty prompt. Initial work by \citet{hendel2023incontext} demonstrates the efficacy of task vectors for steering model behavior in domains including translation, semantic knowledge (country-capital mappings) and syntactic rules. \citet{todd2024functionvectorslargelanguage} extends this general paradigm to recover finer-grained representations from the activations of individual attention heads. Subsequent work has since applied task vectors to causal interventions in a variety of applications from visual image-masking tasks \citep{hojel2024findingvisualtaskvectors} to more naturalistic domains \citep{kang2025adaptivetaskvectorslarge}, while improvements to the paradigm itself include training mechanism modifications to induce task vectors \citep{yang2025taskvectorsincontextlearning} and the superposition of multiple task vectors during in-context learning \citep{xiong2024oncellmsincontextlearn}.

\paragraph{Difference-in-means (DiffMean)}
While task vectors demonstrate, in-practice, that transformer hidden layers encode abstract rules for task representations, these representations are often noisy as a result of extracting the \textit{entire} hidden state at a given token position \citep{hendel2023incontext}. \citet{marks2024geometry} introduce a fine-grained method for extracting the steering vector as a linear direction in the activation space that represents the difference between the mean hidden states of positive and negative prompt examples. This vector, represented as $\theta$, consistently shifts model activations along the direction of the difference, from positive to negative or negative to positive (where \textit{negative} and \textit{positive} are examples along some dimension). \citet{marks2024geometry} demonstrate that these vectors capture significantly less noise compared to simple linear probes, and generalize to a range of naturalistic domains.

\paragraph{Sparse autoencoders}
A key difficulty in extracting steering directions from model activations is the dense nature of these activations: a single neuron will activate for many different tasks and concepts as a part of the distributions representing those concepts. \textit{Sparse autoencoders} attempt to solve this issue by learning a compressed set of activations $\mathbf{z}$ corresponding to the activations in a given transformer layer $l$, with a sparsity penalty applied $\mathcal{L}_{\text{sparse}}$ \citep{cunningham2023sparse}. This allows for the unsupervised discovery of human-interpretable concepts, represented by the activations of single SAE neurons. Formally, the SAE objective is 
\[
\mathcal{L} = \lVert \mathbf{h}^{(l)} - D(E(\mathbf{h}^{(l)})) \rVert_2^2 + \lambda \, \mathcal{L}_{\text{sparse}},
\]
where $\mathbf{h}^{(l)}$ are the transformer activations, $E$ and $D$ are the encoder and decoder, and $\lambda$ is the sparsity penalty. The concepts identified by SAEs range from world knowledge and semantic properties to syntactic features and compositional abstractions \citep{shu2025surveysparseautoencodersinterpreting}. Interestingly, SAE neurons can be used to steer model outputs along dimensions of interest, such as translating LLM outputs to French given the activation of a corresponding latent feature in the sparse autoencoder \citep{he2025saifsparseautoencoderframework}. While the extent to which SAEs provide a practical means of model steering and interpretability is disputed \citep{wu2025axbench}, subsequent work provides methods for obtaining more interpretable, coherent concepts \citep{bussmann2025learningmultilevelfeaturesmatryoshka,rajamanoharan2024jumpingaheadimprovingreconstruction}.
\begin{figure}[t!]
  \centering
  \includegraphics[width=\linewidth]{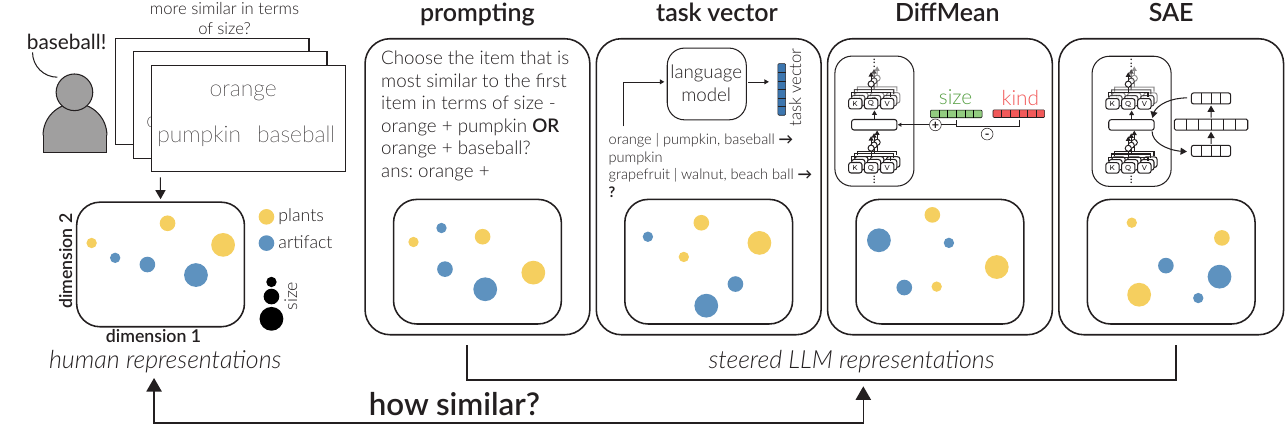}
  \caption{General workflow for deriving human and steered model embeddings.}
  \label{fig:method}
\end{figure}

\subsection{Performance versus alignment}

Task performance and representational alignment are distinct axes of evaluation. \citet{linsley2023performance} found that higher object recognition accuracy led to poorer alignment with human visual cortex. \citet{liu2023aligning} reported tradeoffs between alignment and task performance. This suggests that models can match human performance but rely on different internal representations \citep{firestone2020performance}. As \citet{khamassi2024strong} argue, evaluations should assess not only what systems do, but how. Thus, we evaluate both competence and alignment.

\section{Methods}

\subsection{Dataset}

We use the Round Things Dataset \citep{giallanza2024integrated} to evaluate these aspects. This dataset comprises 46 concrete objects: 25 human-made artifacts and 21 natural kinds (fruits and vegetables). All items are roughly spherical in shape, varying along a discrete \textit{kind} dimension (artifacts versus plants) and a continuous \textit{size} dimension. The clear, unambiguous ground truth nature of the underlying dimensions governing size and kind, coupled with the fact that these are dimensions people consider in everyday human life make this a uniquely well scoped yet ecologically relevant dataset for assessing the effects of steering in language models. We describe the dataset in further detail below.

\paragraph{Dataset design.} The Round Things Dataset was specifically designed to investigate how semantic representations can be contextually modulated along orthogonal dimensions. The choice of roughly spherical objects serves multiple methodological purposes: (1) it facilitates size comparisons by minimizing shape-based confounds, (2) it ensures that size judgments are based primarily on scale rather than geometric complexity, and (3) it creates a controlled stimulus set where the two key dimensions of interest---kind and size---are orthogonal rather than confounded.

The dataset was constructed to include commonly known objects to ensure high familiarity across participants. Each item possesses an empirically-derived ground-truth average diameter obtained through systematic internet searches. Specifically, the authors conducted Google searches using the phrase ``average diameter of a \_\_\_\_\_,'' recording the answer yielded by search engine summaries. When searches returned size ranges, the midpoint was taken as the median diameter. For items that naturally occur in multiple sizes (e.g., pumpkins, yoga balls), searches were refined by adding ``medium-size'' to obtain more standardized measurements.

\paragraph{Data composition.} The 46 items are distributed across two semantic categories: 25 human-made artifacts (such as balls, spherical tools, and round household objects) and 21 fruits and vegetables (such as apples, oranges, and other roughly spherical produce). Importantly, object sizes are approximately balanced across the two categories, ensuring that neither semantic domain is systematically associated with larger or smaller items. This balanced distribution is crucial for investigating how semantic category membership interacts with size-based similarity judgments without confounding effects.

The dataset thus provides stimuli that differ along two clear but unrelated abstract semantic dimensions: the discrete taxonomic dimension of kind (artifact versus natural) and the continuous physical dimension of size. This orthogonal structure allows for controlled investigation of how task context can selectively emphasize one dimension while de-emphasizing another, and whether such emphasis preserves or eliminates information from the non-target dimension.

Human behavioral data collected using this dataset has been used to derive two-dimensional embeddings via crowd-kernel ordinal embedding algorithms, allowing for quantitative analysis of how different task contexts reshape the geometric organization of semantic space while preserving cross-context similarities.

\subsection{Triadic similarity judgment task}

We employ a triadic judgment task where both humans and LLMs identify which of two items $x_1$ or $x_2$ is most similar to a reference item $x_{\text{ref}}$ along a specified semantic dimension. Participants are instructed to judge which of the two options is most similar to the target item either ``in terms of size'' or as ``a more similar kind of thing.''

This paradigm is motivated by the hypothesis that when people estimate similarity along specific dimensions, they use task instructions to steer their semantic representations toward task-relevant features. We test whether different LLM steering methods can similarly modulate representations to achieve both high accuracy (competence) and human-like similarity patterns (alignment).

The dataset's balanced design—with sizes approximately distributed across both semantic categories—enables investigation of how task instructions influence the relative weighting of size versus kind information in similarity computations. This provides insights into whether context-dependent steering preserves or eliminates information from non-target dimensions, revealing the flexibility and constraints of semantic representations under controlled processing conditions.

\subsection{Steering methods}

We evaluated four different steering methods---prompting, task vectors, DiffMean, and SAEs---alongside two prompting baselines (zero-shot and in-context) for comparison. All methods were tested on open-source LLMs capable of running on a single NVIDIA H100 GPU. Figure~\ref{fig:method} shows the general workflow for applying these steering methods to the triadic judgment task.

For the prompting-based approaches, different steering conditions required tailored prompts to elicit appropriate similarity judgments along specific semantic dimensions. Each prompt was designed to direct the model's attention toward either kind-based or size-based similarity, or to allow for general similarity assessment in the neutral condition, as shown in Table~\ref{tab:prompts}. All trials received independent prompt evaluation to ensure consistent application of each steering method.

For the activation-based steering methods (task vectors, DiffMean, and SAEs), we applied the corresponding interventions to the model's internal representations while using a consistent neutral prompt format across all trials. This approach allows us to isolate the effects of representational steering from prompt-based steering, providing a comprehensive comparison of different steering mechanisms.

Detailed implementation specifics for each steering method are provided in Appendix~\ref{sec:app_steering}.

\begin{table}[t]
\centering
\small
\renewcommand{\arraystretch}{1.2}
\caption{Prompt templates used for different steering conditions. The placeholder \{p\} is replaced with the triplet comparison question: "Which item is most similar to \{item\_x\}: \{item\_y\} or \{item\_z\}?"}\label{tab:prompts}
\begin{tabular}{p{2.5cm}p{10.5cm}}
\toprule
\textbf{Condition} & \textbf{Prompt Template} \\
\midrule
Kind & \texttt{Choose the second item that is most similar to the first item in terms of the KIND of thing it is. Respond only with the name of the item exactly as written. \{p\}} \\[0.3cm]

Size & \texttt{Choose the second item that is most similar to the first item in terms of SIZE. Respond only with the name of the item exactly as written. \{p\}} \\[0.3cm]

Neutral & \texttt{Choose the second item that is most similar to the first item. Respond only with the name of the item exactly as written. \{p\}} \\
\bottomrule
\end{tabular}
\end{table}

\subsection{Analysis Approach}

\paragraph{Embedding algorithm.} We applied an ordinal embedding algorithm \citep{tamuz2011adaptively, sievert2023efficiently} to similarity judgments from both humans and models, creating semantic spaces where two items are proximal when they were often judged as similar to one another relative to a third item. The embedding algorithm minimized the crowd-kernel triplet loss \citep{tamuz2011adaptively} by optimizing Euclidean distances between word pairs. To ensure reliable estimates, we reserved 20\% of our data for validation purposes and monitored crowd-kernel loss throughout the fitting procedure.

Under this framework, each triplet judgment yields an ordinal constraint of the form "item $i$ is closer to item $j$ than to item $k$." Let $D^\star$ be the squared Euclidean distance matrix of unknown embedding points $\{x_i^\star\}_{i=1}^n\subset\mathbb{R}^d$. The ordinal embedding model assumes a monotone link function (e.g., logistic):
\begin{equation}
\Pr\big[y_{(i;j,k)}=1\big]~=~f\!\big(D^\star_{ik}-D^\star_{ij}\big),
\qquad f(0)=\tfrac12.
\label{eq:link}
\end{equation}
The ordinal embedding algorithm minimizes an empirical surrogate of the negative log-likelihood using crowd-kernel triplet loss over a low-rank parameterization.

\textit{Sample complexity considerations.} If the true embedding has rank $d$ and triplet judgments are sampled approximately uniformly at random, then with high probability the out-of-sample prediction error is
\begin{equation}
\mathcal{E}_{\mathrm{pred}}~=~\tilde O\!\left(\sqrt{\frac{d\,n\log n}{|S|}}\right),
\qquad
\Rightarrow\quad
|S|~=~\tilde \Theta\!\big(d\,n\log n\big)
\label{eq:dnlogn}
\end{equation}
Thus, $\tilde \Theta(nd\log n)$ triplet judgments suffice for accurate prediction of new comparisons, and at least $\Omega(d\,n\log n)$ ordinal comparisons are information-theoretically necessary \citep{jain2016finitesamplepredictionrecovery}. For our implementation with $n=46$ items and $d=2$ dimensions, we collected $N_{\mathrm{triplets}} = c \cdot d \cdot n \log n$ triplet judgments per steering method (where $c \approx 5$ based on our 2,500 judgments), balancing statistical efficiency with computational constraints.

\paragraph{Accuracy measurement.} We evaluated the competence of each steering method by measuring LLM accuracy on the triadic judgment task compared to ground-truth and human performance. Human accuracy benchmarks were derived from human embeddings by evaluating the same 2,500 triplets used in model assessments: when the Euclidean distance from $x_{\text{ref}}$ to $x_1$ was shorter than from $x_{\text{ref}}$ to $x_2$, we classified $x_1$ as the correctly identified more similar item. This approach enables direct comparison between human and model performance on identical triplet sets, providing a unified measure of task competence across all steering conditions.

\paragraph{Alignment analysis.} To quantify alignment between human and LLM representations, our main measure of human-model alignment was the amount of variance in human semantic embeddings explained by model-based semantic embeddings ($R^2$) after Procrustes aligning the two spaces \citep{gower1975generalized}. Procrustes transformation aligns two vector spaces of equal dimensionality by finding the optimal set of linear transformations (rotations, scalings, and translations) to minimize the sum of squared distances between corresponding points in the two spaces (residual SSE). Using the sum of squared distances in the human embeddings (human SSE) as the target, we computed an $R^2$ metric as:

$$R^2 = 1 - \frac{\text{residual SSE}}{\text{human SSE}}$$

This approach measures how well variations in pairwise distances in LLM-derived representations correspond to those in human representations after allowing for optimal linear transformations in the representational geometry. Since both embedding spaces are of the same dimensionality and permitting these linear transformations makes our alignment estimate maximally generous, this provides a robust upper bound on the correspondence between human and model semantic representations.

\begin{figure}[ht!]
  \centering
  \includegraphics[width=\linewidth]{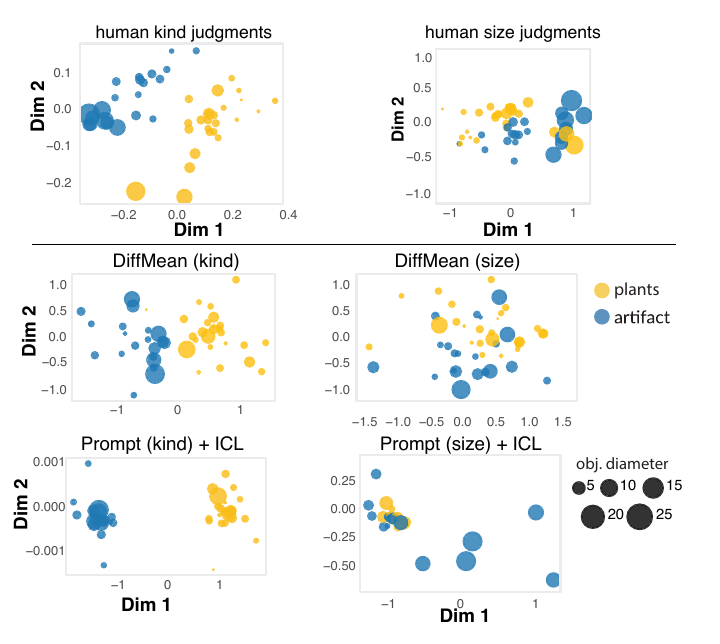}
  \caption{Representational geometry of the concepts in the Round Things Dataset based on embeddings derived from triadic judgments for humans (top row) and \texttt{gemma-2-27b-it} (bottom 2 rows) using two steering methods. Plots for all methods can be seen in \ref{sec:full_embedding_plots}}
\label{fig:embeddings}
\vspace{-2em}
\end{figure}

\section{Results}

We assessed whether steering led to \textit{correct judgments} (competence) and how the steering influenced the underlying \textit{representational geometry} in models with respect to that of humans (alignment). As noted, we evaluated three models of varying sizes across two model families: \texttt{gemma-2-9b-it}, \texttt{gemma-2-27b-it}, and \texttt{qwen-2.5-14b-it}.
We found that the main findings were broadly consistent across these models.


\subsection{Model representations show default kind alignment}

\begin{figure}[t!]
    \centering
    \includegraphics[width=.8\textwidth]{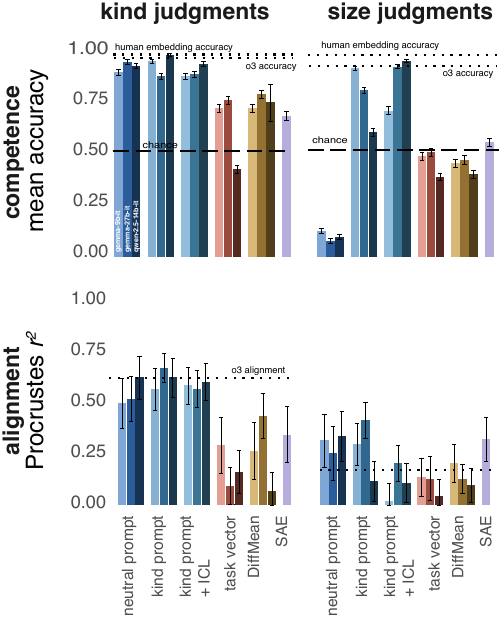}
  \caption{Steering accuracy (top row) and alignment of steered LLM representations to human representations (bottom row) for each steering technique. the dashed line labeled `human embeddings' corresponds to how accurately human judgments can be predicted from human embeddings. SAE results are reported only for \texttt{gemma-9b-it} due to compute limitations.}
  \label{fig:results}
\end{figure}

Figure~\ref{fig:results} shows how successful different steering methods were in guiding LLMs to make accurate kind and size judgments.
Models were generally biased towards `kind' representations when prompted with neutral prompts, which simply asked LLMs to indicate which of the two options was most similar to the target without additional context.
Two lines of evidence supported this.
First, models were significantly more accurate at the kind judgments than the size judgments under the neutral prompt (\( \beta_{\text{kind}} = 4.16\), \( p < 0.001 \), \( \beta_{\text{size}} = 0.05\), \( p = 0.86 \) for \texttt{gemma-2-9b-it}; \( \beta_{\text{kind}} = 6.09\), \( p < 0.001 \), \( \beta_{\text{size}} = 0.23\), \( p = 0.19 \) for \texttt{qwen-2.5-14b-it};).
Second, model representations were more aligned to `kind' \textit{human} representations under the neutral prompt (bottom row of Figure~\ref{fig:results}): While neutral prompt embeddings were strongly aligned with human kind embeddings, alignment was weaker for size embeddings. For the kind dimension, the 27B, 9B, and 14B variants correlated highly with human judgments (\( r_{27\mathrm{B},\text{kind}} = 0.724,\, p < .005;\; r_{9\mathrm{B},\text{kind}} = 0.708,\, p < .005;\; r_{14\mathrm{B},\text{kind}} = 0.793,\, p < .001\)). Correlations with human size embeddings were comparatively weaker (\( r_{27\mathrm{B},\text{size}} = 0.495,\, p = 0.030;\; r_{9\mathrm{B},\text{size}} = 0.553,\, p = 0.013;\; r_{14\mathrm{B},\text{size}} = 0.570,\, p = 0.010 \)). All \( p \)-values comparing correlation coefficients were computed by first computing a Fisher \( r \rightarrow z \) transform and then conducting the appropriate $t$-test.

\subsection{Prompting outperforms other steering methods at predicting ground truth and human alignment}

Intervention methods (SAEs, task vectors, and DiffMean) consistently performed worse than prompting in producing correct (ground-truth) judgments for both the \textit{kind} (\( \chi^2(1) = 952.03,\ \beta = 3.06,\ p < 0.001 \) for \texttt{gemma-2-27b-it}; \( \chi^2(1) = 1243.92,\ \beta = 2.06,\ p < 0.001 \) for \texttt{gemma-2-9b-it}; \( \chi^2(1) = 1831.93,\ \beta = 2.99,\ p < 0.001 \) for \texttt{qwen-2.5-14b-it}) and \textit{size} conditions (\( \chi^2(1) = 2055.36,\ \beta = 2.34,\ p < 0.001 \) for \texttt{gemma-2-27b-it}; \( \chi^2(1) = 2818.67,\ \beta = 2.69,\ p < 0.001 \) for \texttt{gemma-2-9b-it}; \( \chi^2(1) = 805.76,\ \beta = 1.31,\ p < 0.001 \) for \texttt{qwen-2.5-14b-it}). In the \textit{kind} condition, task vectors and DiffMean did not differ reliably in accuracy for \texttt{gemma-2-9b-it} (\( \chi^2 = 0.17,\ p = 0.68 \)), though DiffMean outperformed task vectors for \texttt{gemma-2-27b-it} (\( \chi^2 = 5.73,\ p = 0.017 \)) and for \texttt{qwen-2.5-14b-it} (\( \chi^2 = 16.20,\ p < 0.001 \)), while in the \textit{size} condition we observed a small but reliable benefit in performance for task vectors over DiffMean (\( \chi^2 = 4.47,\ p = 0.035 \) for \texttt{gemma-2-27b-it}; \( \chi^2 = 5.64,\ p = 0.018 \) for \texttt{gemma-2-9b-it}; \( \chi^2 = 18.75,\ p < 0.001 \) for \texttt{qwen-2.5-14b-it}). SAEs were more effective at predicting the ground truth in the \textit{size} condition relative to DiffMean and task vectors (\( \chi^2 = 26.24{-}55.83,\ p < 0.001 \)), but in the \textit{kind} condition were reliably worse than the prompting methods (\( \chi^2 = 314.81{-}460.96,\ p < 0.001 \)) and did not differ reliably from the other steering methods (\( \chi^2 = 1.48{-}2.68,\ p = 0.10{-}0.22 \)).


In the \textit{kind} condition, prompting methods were also most effective in aligning model representations with humans. A regression model predicting representational alignment from steering methods revealed a significant improvement for prompting over non-prompting methods (\( \beta = 0.23,\ t = 4.09,\ p = 0.003 \); including \texttt{qwen-2.5-14b-it}, \( \beta = 0.31,\ t = 4.99,\ p < 0.001 \)), consistent with the large Procrustes correlations observed for all prompt-based methods (\( r_{\text{prompt–kind–27b}} = 0.82,\ p < 0.001;\; r_{\text{prompt–kind–9b}} = 0.76,\ p < 0.001;\; r_{\text{prompt–kind–14b}} = 0.79,\ p < 0.001;\; r_{\text{prompt–ICL–27b}} = 0.76,\ p < 0.001;\; r_{\text{prompt–ICL–9b}} = 0.77,\ p < 0.001;\; r_{\text{prompt–ICL–14b}} = 0.78,\ p < 0.001 \)). The other steering methods showed weaker, but in several cases still significant, alignment (\( r_{\text{TV–9b}} = 0.54,\ p = 0.015;\; r_{\text{TV–14b}} = 0.41,\ p = 0.085;\; r_{\text{DM–27b}} = 0.66,\ p = 0.001;\; r_{\text{DM–9b}} = 0.52,\ p = 0.022;\; r_{\text{DM–14b}} = 0.26,\ p = 0.285;\; r_{\text{SAE–9b}} = 0.59,\ p = 0.007 \)).

In the \textit{size} condition, we observed no such benefit in the prompting conditions. A regression showed no prompting advantage (\( \beta = 0.034,\ t = 0.43,\ p = 0.674 \)), consistent with comparatively weaker Procrustes correlations for zero-shot and in-context prompts relative to the kind condition (\( r = 0.633,\ p = 0.0028 \) for 27B; \( r = 0.536,\ p = 0.017 \) for 9B; \( r = 0.337,\ p = 0.161 \) for 14B; prompt–ICL \( r = 0.444,\ p = 0.056 \) for 27B, \( r = 0.142,\ p = 0.567 \) for 9B, and \( r = 0.325,\ p = 0.177 \) for 14B). Among the intervention methods, only the SAE condition produced significantly aligned representations (\( r_{\text{SAE–9b}} = 0.558,\ p = 0.012 \)).


While generally concordant with other work that highlights the importance of prompt-based steering, the present work distinguishes itself in focusing on measures of \textit{representational alignment }\citep{sucholutsky2023getting} -- that is, how the internal representations in any two agents are aligned.
While these findings are helpful in showing that prompting, a largely accessible and simple method, outperforms more bespoke methods, it also highlights the difficulty in guiding model behavior through intervening on model internals -- a key goal of mechanistic interpretability research. We discuss this further in the following section.

\section{Conclusion}

We evaluated a set of popular LLM steering techniques using a well-established method in the cognitive sciences---triadic similarity judgments. By using both steered LLM and human judgments as input into embedding algorithms that capture an agent's representational geometry, we applied a commensurate standard to evaluate how human-like different steering methods are. Critically, we looked beyond raw accuracy (competence) and evaluated steered models on how strongly their embeddings aligned with human embeddings. Similar to prior work \citep{wu2025axbench}, we found that prompting methods tended to outperform other steering techniques in terms of both accuracy and alignment. We also found that LLMs, without any steering, tended to privilege the \textit{kind} dimension of similarity over the \textit{size} dimension, indicated both by the stronger alignment of the neutral prompt condition with embeddings from human kind judgments and by the difficulty steered models had aligning with human size representations.

Our work furthers a critical distinction between model interpretability  and model controllability. Methods that give limited controllability might uncover model representations and components that do not play \textit{causal} roles in driving model behavior. 
As it currently stands, we find limited evidence for non-prompting methods successfully steering model representations to align with human behavior or representations for a relatively simple class of similarity judgments.
While models might continue to set record-breaking benchmark performances, there is no guarantee they will begin to align with human minds.
However, methods for explaining the success and failures of certain methods over others will continue to be important for understanding how LLM representations are transformed by various forms of `control.'
In addition to the methods explored here, rational analysis of language models through cognitive science-inspired methods has begun to shed light on other aspects  of LLM behavior, including chain-of-thought prompting, in-context learning, and the effects of pretraining \citep{prystawski2023think, wurgaft2025context, mccoy2024embers}.
Continued efforts on this front may also help uncover the levers necessary for steering of models in human-like ways. 

Taken together, these findings provide a novel perspective on how steering methods can and should be evaluated in order to assess how aligned steered model representations are to those of people. We find converging evidence that prompt-based steering is currently the best route for both accurate and aligned steering, and that some axes such as kind are privileged over size. Future work should seek to further integrate insights from controlled semantic cognition \citep{giallanza2024integrated} to uncover the basis of prompting's success in guiding LLMs' learned representations in a context-sensitive manner, and to test a wider variety of contexts beyond size and kind judgments.

\section{Limitations}

\textbf{Model suite}. Due to compute constraints we only evaluated models that could be run on consumer-grade GPUs (\texttt{gemma-2-9b-it}, \texttt{gemma-2-27b-it}, and \texttt{qwen-2.5-14b-it}). Further, we did not evaluate \texttt{gemma-2-27b-it} on the SAE method due to the lack of available trained SAEs for that model. Despite this, we believe the presented results constitute a fair comparison between methods. Future work can scale up this approach to larger models. The efficacy of different steering methods for both accuracy and alignment may scale differently with model parameter size. Between the 9B--27B scale we did not observe drastic changes, but this could change at larger scales.

\textbf{Method suite}. While we evaluated a representative set of steering methods for `online' steering based on prompts or activation changes, we did not exhaustively test alternative `in-weight' methods of changing model behavior \citep{anand2024dual}, including supervised fine-tuning (SFT), low-rank adaptation (LoRA), and others. Since we are modeling human steering, which is thought to invoke online `control' processes as opposed to in-weight learning \citep{giallanza2024integrated}, we argue that our chosen set of methods provided fair comparisons to the human data. Nevertheless, future work should compare the relative efficacy of in-weight vs.\ in-context styles of steering.

\textbf{Order effects}. While we randomized the order in which concepts were presented to the models for the triplet task---a common method in the behavioral sciences to protect against order effects---we note that we only ran the LLM evaluations with a single ordering. Since humans were also only presented with a single overall ordering, and our goal was to make commensurate comparisons across agent types, we chose to trade off robustness against order effects for fairer comparisons.

\textbf{Limited evaluation methods}. We relied on the triadic similarity judgment task since it is a well-established and validated method for evaluating human mental representations. In practice, there are more complex and naturalistic behaviors that humans perform that require `steering' semantic representations \citep{giallanza2024integrated}. In the future, it will be crucial to incorporate such tasks when benchmarking different language model steering methods.

\textbf{Acknowledgments.} We thank members of the Knowledge and Concepts Lab and Lupyan Lab for helpful feedback. We also thank the Center for High Throughput Computing for providing the compute resources for this project. 

\textbf{Code and materials: }\url{https://github.com/Knowledge-and-Concepts-Lab/SAE_concepts}




\bibliography{colm2026_conference}
\bibliographystyle{colm2026_conference}
\clearpage

\appendix
\section{Steering method and embedding method implementations}
\label{sec:appendix}

\subsection{Detailed implementation of steering methods}\label{sec:app_steering}

Let $f$ be a decoder-only language model with $L$ layers and hidden size $d$. Each triplet comparison is denoted as $t_i = (x_{\text{ref},i}, x_{1,i}, x_{2,i})$ for $i = 1, \dots, n$. Each prompt consists of a sequence of $n$ triplets $[t_1, \dots, t_n]$, serialized into a token sequence $x = [x_1, \dots, x_T]$. The model produces hidden states $h_j^l \in \mathbb{R}^d$ at each token position $x_j$ and layer $l$.

For all prompting, prompts are formatted as natural language strings of the form:

\begin{quote}
\texttt{Choose the item that is most similar to the first item in terms of \textit{<d>}. Respond only with the name of the item exactly as written.}\\
\texttt{\textit{<x\_ref>} + \textit{<x\_1>} OR \textit{<x\_ref>} + \textit{<x\_2>}?}\\
\texttt{answer: \textit{<x\_ref>} +}
\end{quote}

For steering methods, training examples consist of triplets without a natural language instruction like so:

\texttt{\textit{<x\_ref>} + \textit{<x\_1>} OR \textit{<x\_ref>} + \textit{<x\_2>}?}\\
\texttt{\textit{<x\_ref>} + \textit{<x\_answer>}}

Where examples are concatenated by commas. Each training example consisting of $n$ triplets has a final incomplete training instance like so:

\texttt{\textit{<x\_ref>} + \textit{<x\_1>} OR \textit{<x\_ref>} + \textit{<x\_2>}?}\\
\texttt{\textit{<x\_ref>} +}

This "+" token is used to extract and steer representations for a corresponding "+" token in the zero-shot test example, which is identical to the final training example:

\texttt{\textit{<x\_ref>} + \textit{<x\_1>} OR \textit{<x\_ref>} + \textit{<x\_2>}?}\\
\texttt{\textit{<x\_ref>} +}

Fields are interpolated for some $d \in \{\textit{size}, \textit{kind}, \textit{neutral}\}$, triplet $t_n = (x_{\text{ref}}, x_1, x_2)$, and answer $t_{answer}$ given the dimension $d$. We extract activations, logits, and apply all steering methods at the final input token $x_T = \texttt{+}$ in the last triplet $t_n$, depending on each method.

\subsection*{Zero-shot Prompt}

In the zero-shot condition, the model is given a single triplet $t_n = (x_{\text{ref}}, x_1, x_2)$ and is asked to make a discrimination along a semantic dimension $d \in \{\textit{size}, \textit{kind}, \textit{neutral}\}$.

\subsubsection*{Prompt with In-Context Examples}

In the in-context condition, the model is given a sequence of $n = 15$ complete triplets $[t_1, \dots, t_{15}]$ and is asked to make a discrimination for the final triplet $t_{15} = (x_{\text{ref}}, x_1, x_2)$ along semantic dimension $d \in \{\textit{size}, \textit{kind}\}$. 

\subsubsection*{Task Vector}

Following \citet{hendel2023incontext}, we extract \textit{task vectors} for the \textsc{kind} and \textsc{size} conditions by first constructing two prompts organized along each condition: \( x_{\text{train}} \), containing 14 complete triplet examples and one final incomplete example (with \texttt{"+"}), and \( x_{\text{test}} \), containing a single zero-shot incomplete triplet.

For each layer \( \ell \in \{0, \dots, L\} \), we extract the hidden activation in the residual stream at the final token position of \( x_{\text{train}} \) (i.e., the \texttt{"+"}) and patch it into the corresponding position in \( x_{\text{test}} \). The language model \( f \) then autoregressively generates a sequence \( x_0, \dots, x_k \) until a complete output is produced.

We repeat this procedure over 200 randomly generated \( (x_{\text{train}}, x_{\text{test}}) \) pairs, selecting the layer \( \ell^*_d \) that yields the highest accuracy. Finally, using this optimal layer \( \ell^*_d \), we repeat the procedure across 2400 additional prompt pairs to generate task vector embeddings for both the \textsc{size} and \textsc{kind} conditions.

\subsubsection*{DiffMean}

\textsc{DiffMean} constructs a steering vector by computing the average difference between latent representations of \textit{positive} and \textit{negative} examples along a target output dimension. Specifically, the mean latent representation of the positive examples is subtracted from that of the negative examples, producing a steering vector. This vector is then \textit{added} (as opposed to task vectors, where it is patched) to the latent representation of a held-out prompt \( x_{\text{test}} \) in order to steer the model's output generation.

For a target steering dimension \( d \in \{\textit{size}, \textit{kind}\} \) and its contrast \( d' \), we generate 15 triplet examples organized along \( d \), and 15 along \( d' \), where the final triplet in each set is incomplete and ends with the \texttt{"+"} token. 

Then, for a given layer \( \ell \in \{0, \dots, L\} \), we extract the residual stream representation \( r_T^\ell \) at the final token position \( T \) from both \( x_{\text{train},d} \) and \( x_{\text{train},d'} \). The \textsc{DiffMean} steering vector is then computed as the difference:

\[
v_{\text{diff}} = r_T^\ell(x_{\text{train},d}) - r_T^\ell(x_{\text{train},d'})
\]

This resulting vector \( v_{\text{diff}} \) defines a direction in the residual stream corresponding to the contrast between the dimensions \( d \in \{\textit{size}, \textit{kind}\} \). 

Similar to the task vector condition, We repeat this procedure over 200 randomly generated \( (x_{\text{train}}, x_{\text{test}}) \) pairs, selecting the layer \( \ell^*_d \) that yields the highest accuracy. Finally, using this optimal layer \( \ell^*_d \), we repeat the procedure across 2400 additional prompt pairs to generate DiffMean embeddings for both the \textsc{size} and \textsc{kind} conditions.

\subsubsection*{SAEs}
We use sparse autoencoders (SAEs) from \textsc{GemmaScope}~\citep{lieberum2024gemmascopeopensparse} to steer the model along interpretable directions in residual space. An SAE is a linear model that decomposes a residual stream vector \( r \in \mathbb{R}^d \) as a sparse linear combination of features, where \( W \in \mathbb{R}^{d \times k} \) is a learned feature dictionary and \( z \in \mathbb{R}^k \) is a sparse activation vector. Each column of \( W \) defines a directional feature in residual space, and only a small number of features are active for any given input.

For each semantic dimension \( d \in \{\textit{size}, \textit{kind}\} \), we construct 20 prompts and identify the feature \( f \in \mathbb{R}^d \) with the highest average activation at layer \( \ell = 20 \). We steer the model by injecting \( c \cdot f \) (with \( c = 50 \)) into the residual stream at layer 20 (the only available layer for \texttt{gemma-2-9b-it} on \textsc{GemmaScope}), and generate 2400 zero-shot completions from held-out prompts \( x_{\text{test}} \). The value of $c$ is selected from a parameter sweep over a subset of $n=250$ prompts. We select the value $c$ which elicits the highest average accuracy for both the kind and size conditions. These completions are used to construct semantic embeddings for each SAE-steered dimension.

\clearpage

\subsection{Procrustes correlations for all prompt and steering methods}\label{sec:procrustes}

The comprehensive set of procrustes correlations for all steering methods, for \texttt{gemma-2-9b-it} and \texttt{gemma-2-27b-it}.

\begin{figure}[ht]
    \centering
    \includegraphics[width=\linewidth]{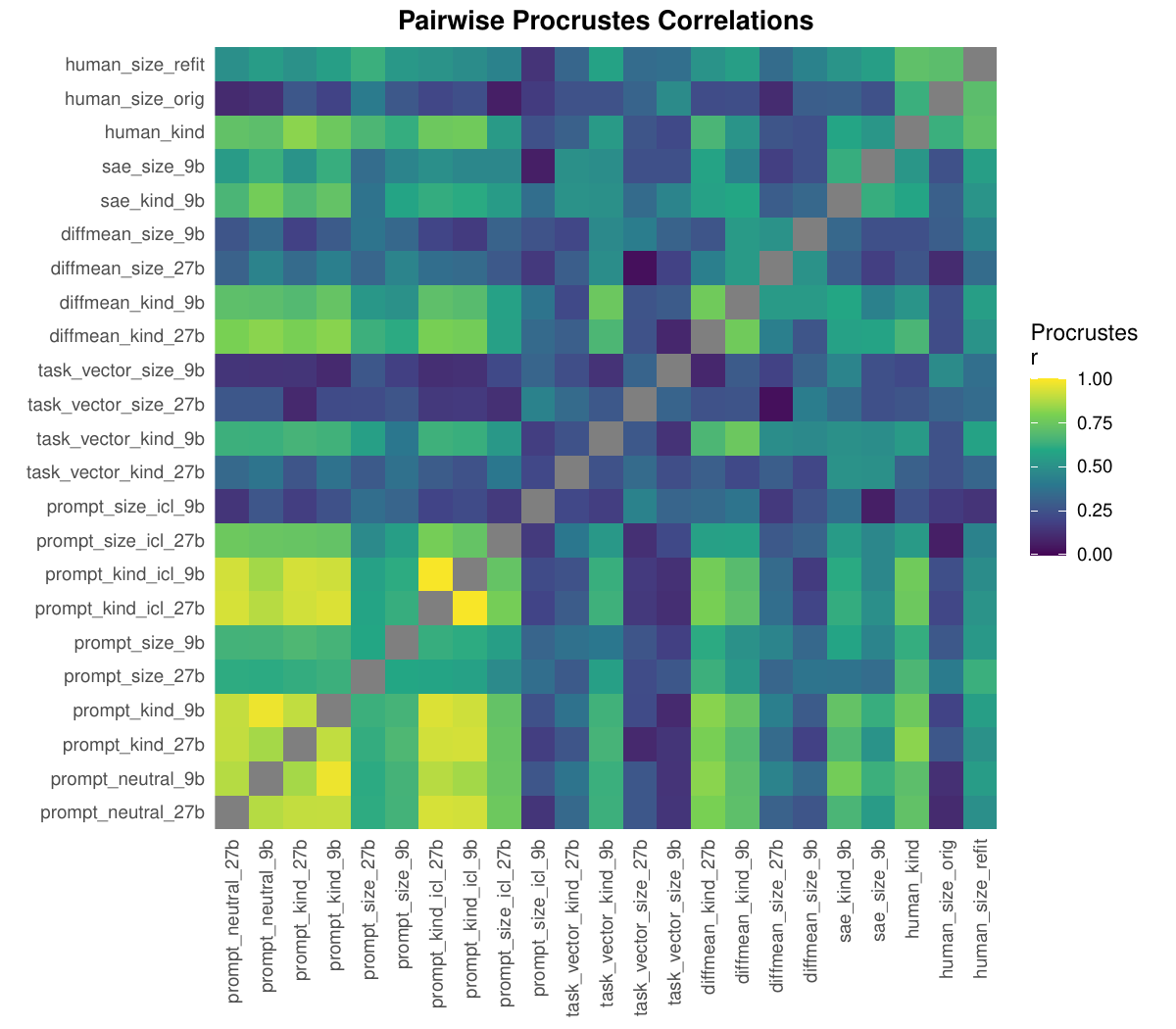}
    \caption{Full procrustes correlations for all methods}
    \label{fig:procrustes}
\end{figure}

\clearpage
\subsection{ICL Prompt Analysis}

We systematically vary the number of example triplet pairs included in the \textsc{kind} condition for \texttt{gemma-2-9b-it} to examine how changes to the input prompt affect model accuracy and human alignment. We find that there is no significant impact of the number of ICL examples on accuracy or alignment. 

\begin{figure}[ht]
    \centering
    \begin{subfigure}[t]{0.48\textwidth}
        \includegraphics[width=\linewidth]{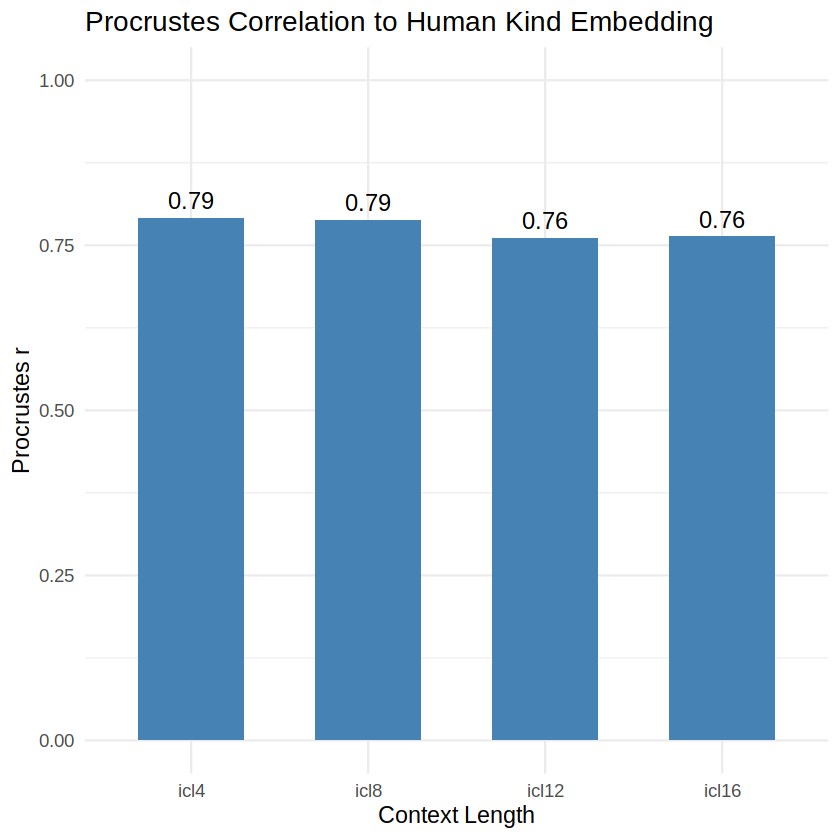}
        \caption{Procrustes correlations with varied ICL examples}
        \label{fig:procrustes_icl}
    \end{subfigure}
    \hfill
    \begin{subfigure}[t]{0.48\textwidth}
        \includegraphics[width=\linewidth]{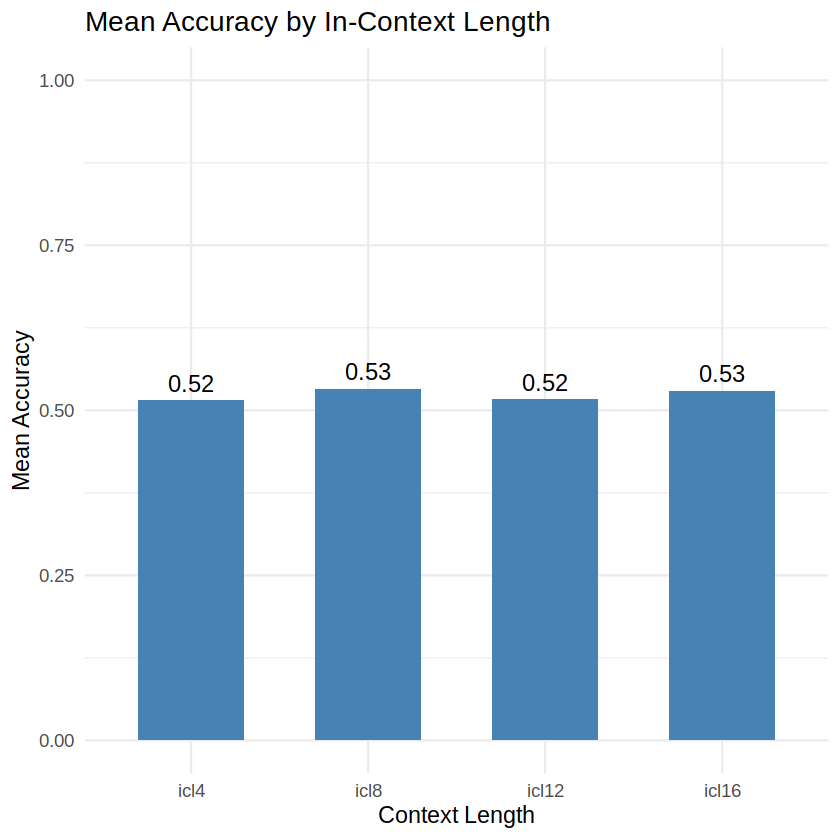}
        \caption{Mean accuracy with varied ICL examples}
        \label{fig:mean_icl}
    \end{subfigure}
    \caption{Impact of varying the number of ICL examples on model performance}
    \label{fig:icl_analysis}
\end{figure}

\subsection{Embeddings Dimensions Analysis}
\label{sec:embeddings_analysis}

Cumulative variance explained by the first $k$ dimensions of the embeddings for \texttt{gemma-2-27b-it}, size condition. The first two dimensions of all embeddings were used for comparisons of representations.

\begin{figure}[ht]
    \centering
    \includegraphics[width=0.8\linewidth]{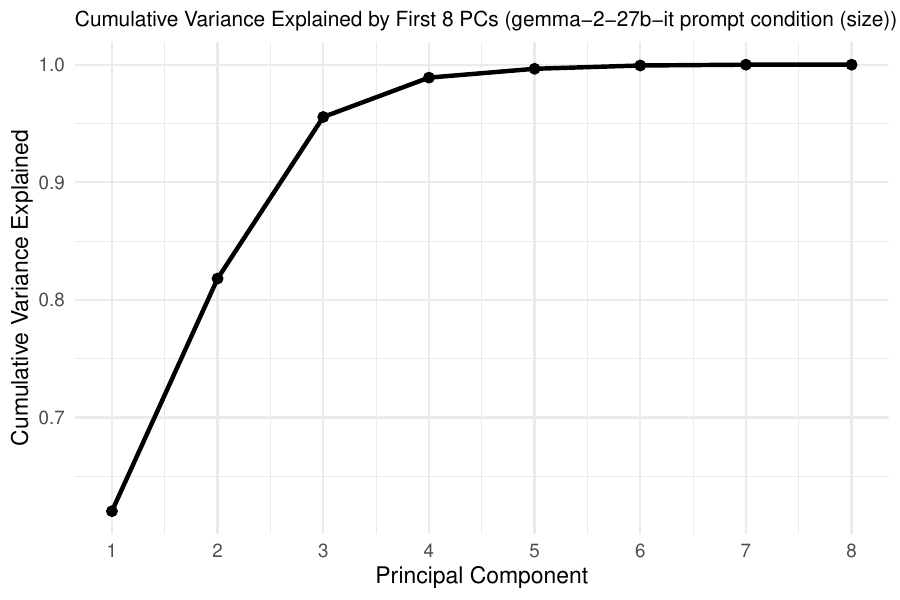}
    \caption{Cumulative variance explained by embedding dimensions}
    \label{fig:cumulative_variance}
\end{figure}



\subsection{Full Embedding Plots}
\label{sec:full_embedding_plots}
\setlength{\fboxsep}{3pt}
\setlength{\fboxrule}{0.5pt}

\begin{figure}[ht!]
    \centering
    \vspace{0.5cm}
    
    \fbox{%
    \begin{subfigure}[t]{0.46\textwidth}
        \centering
        \includegraphics[width=0.95\linewidth]{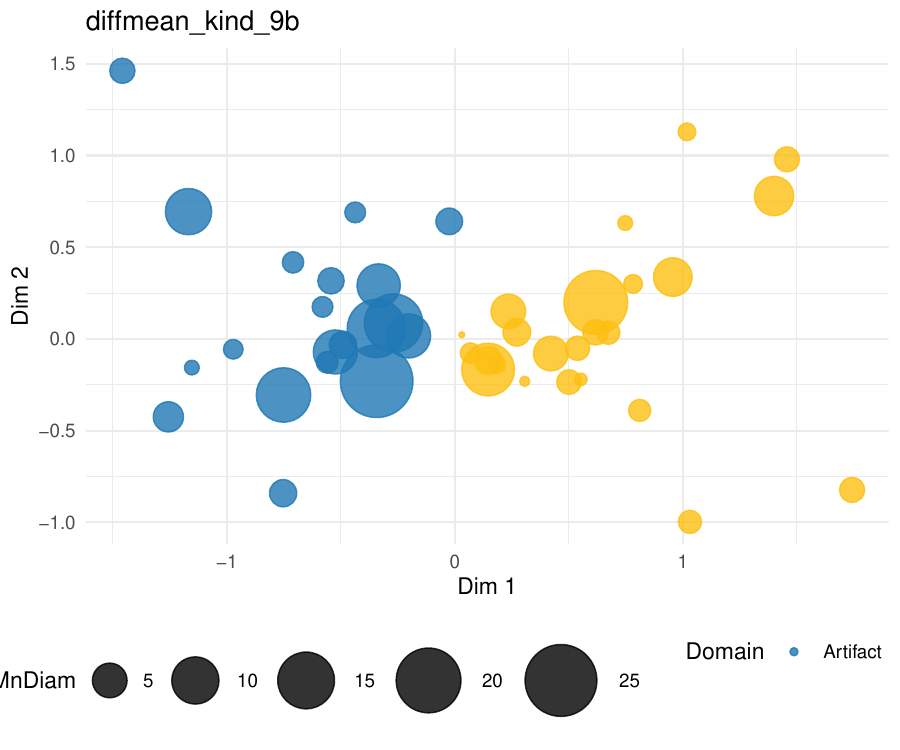}
        \vspace{0.2cm}
        \caption{DiffMean Kind 9B}
        \label{fig:diffmean_kind_9b}
    \end{subfigure}%
    }
    \hfill
    \fbox{%
    \begin{subfigure}[t]{0.46\textwidth}
        \centering
        \includegraphics[width=0.95\linewidth]{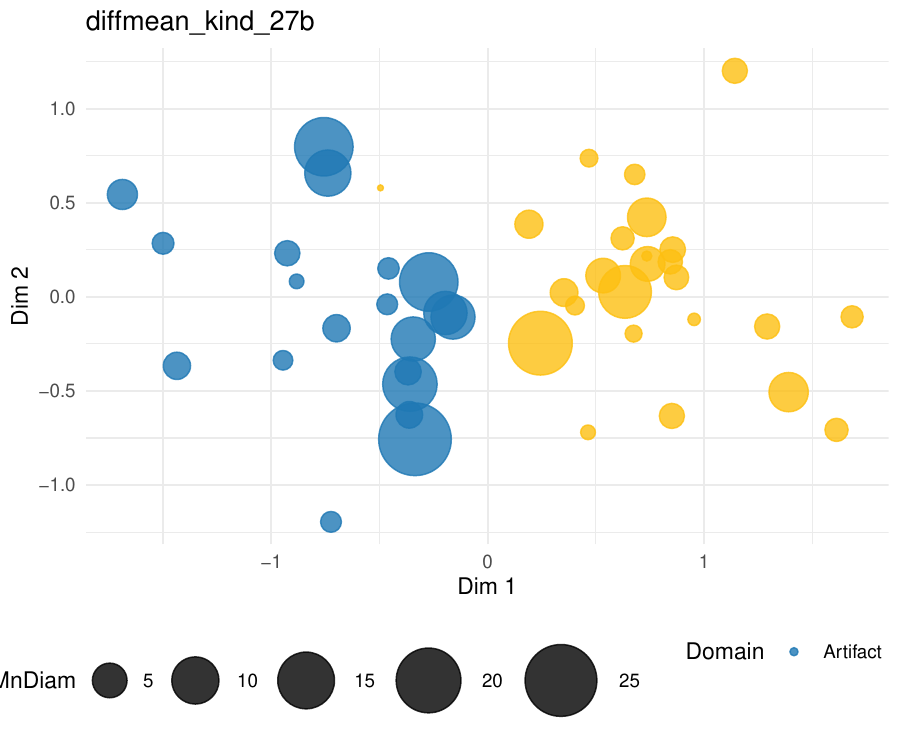}
        \vspace{0.2cm}
        \caption{DiffMean Kind 27B}
        \label{fig:diffmean_kind_27b}
    \end{subfigure}%
    }
    
    \vspace{0.8cm}
    
    \fbox{%
    \begin{subfigure}[t]{0.46\textwidth}
        \centering
        \includegraphics[width=0.95\linewidth]{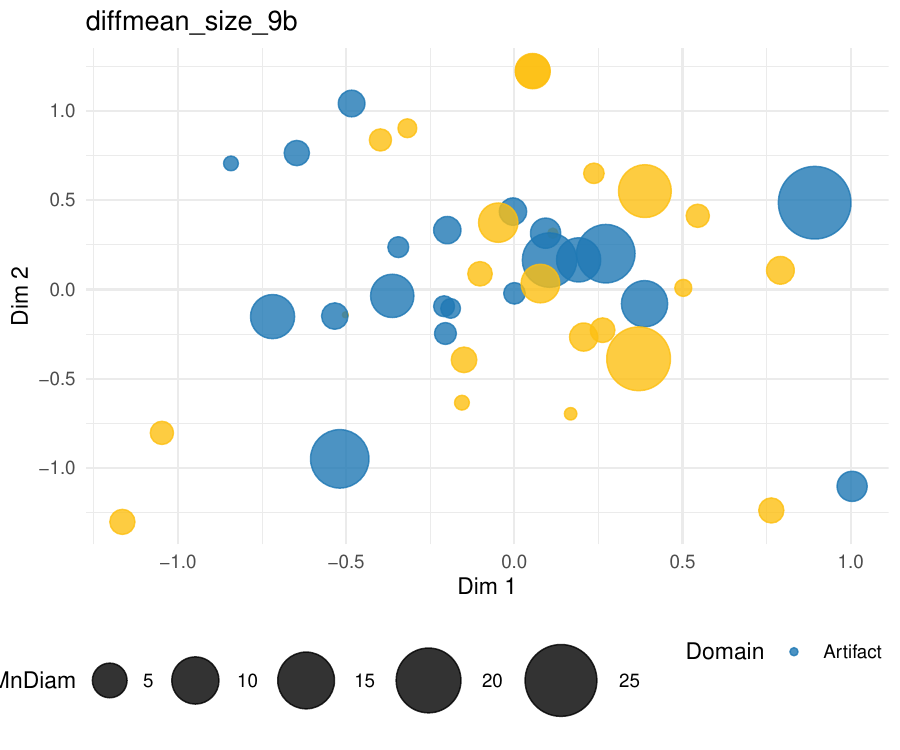}
        \vspace{0.2cm}
        \caption{DiffMean Size 9B}
        \label{fig:diffmean_size_9b}
    \end{subfigure}%
    }
    \hfill
    \fbox{%
    \begin{subfigure}[t]{0.46\textwidth}
        \centering
        \includegraphics[width=0.95\linewidth]{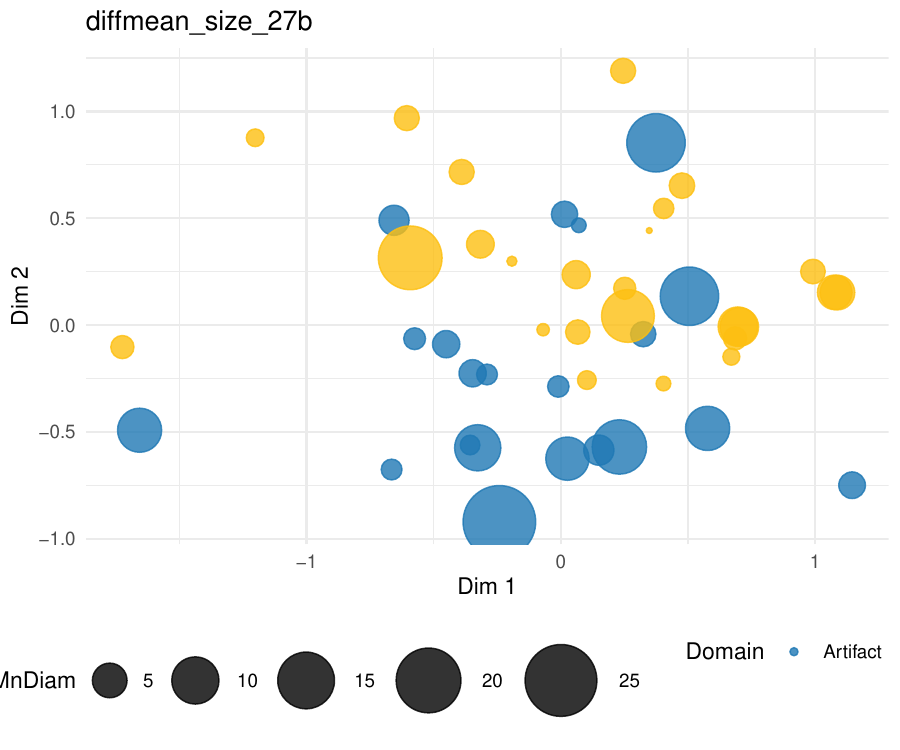}
        \vspace{0.2cm}
        \caption{DiffMean Size 27B}
        \label{fig:diffmean_size_27b}
    \end{subfigure}%
    }
    
    \vspace{0.8cm}
    
    \fbox{%
    \begin{subfigure}[t]{0.46\textwidth}
        \centering
        \includegraphics[width=0.95\linewidth]{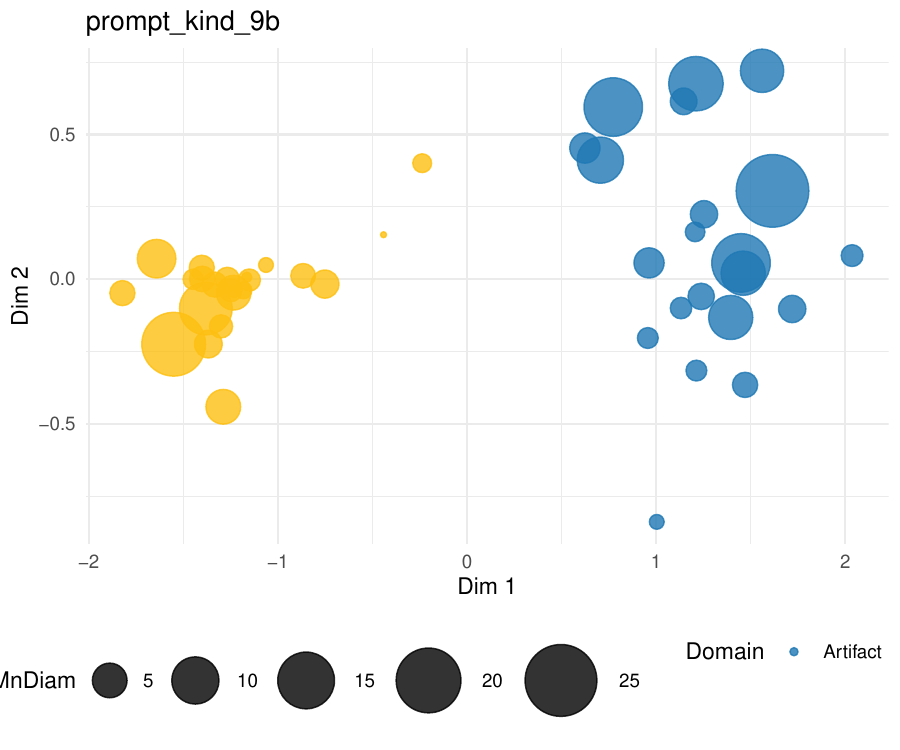}
        \vspace{0.2cm}
        \caption{Prompt Kind 9B}
        \label{fig:prompt_kind_9b}
    \end{subfigure}%
    }
    \hfill
    \fbox{%
    \begin{subfigure}[t]{0.46\textwidth}
        \centering
        \includegraphics[width=0.95\linewidth]{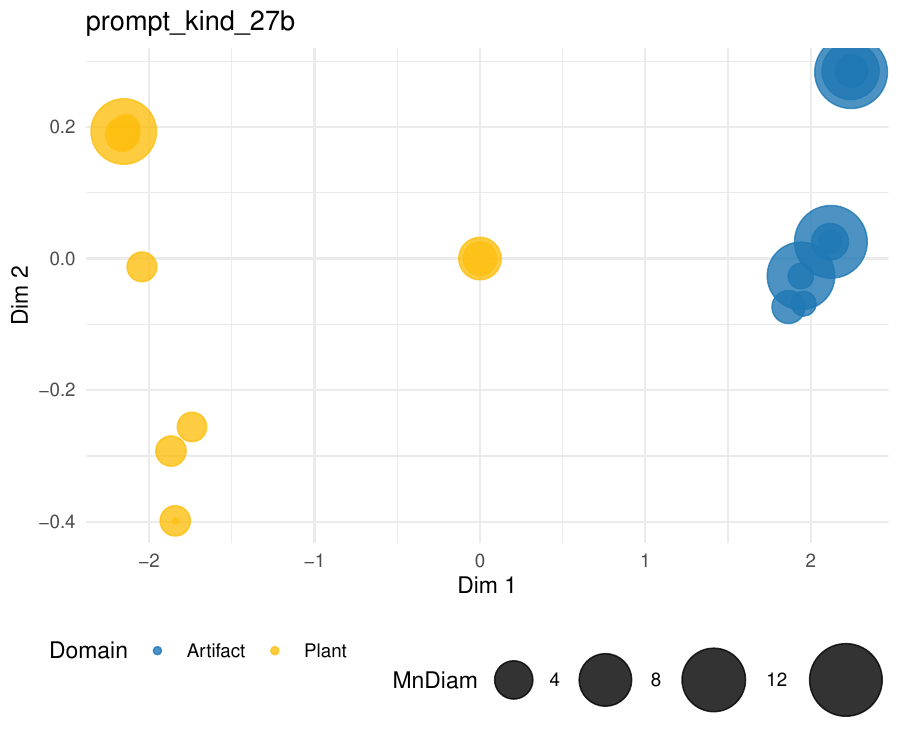}
        \vspace{0.2cm}
        \caption{Prompt Kind 27B}
        \label{fig:prompt_kind_27b}
    \end{subfigure}%
    }
    
    \vspace{0.5cm}
    \caption{Embedding Plots: DiffMean and Prompt Methods}
    \label{fig:embeddings_page1}
\end{figure}

\begin{figure}[p]
    \centering
    \vspace{0.5cm}
    
    \fbox{%
    \begin{subfigure}[t]{0.46\textwidth}
        \centering
        \includegraphics[width=0.95\linewidth]{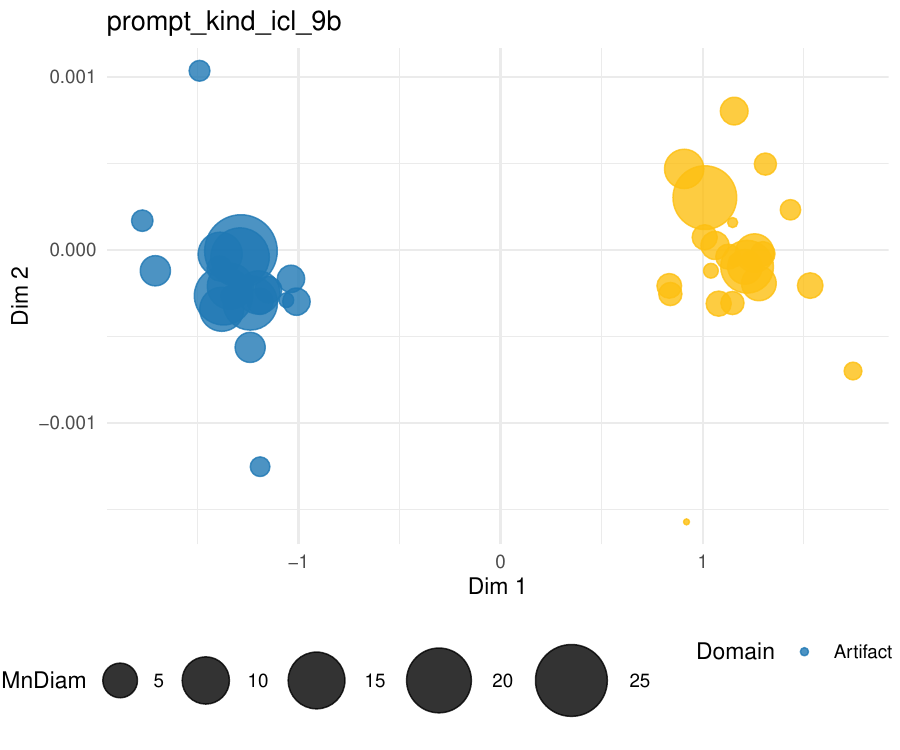}
        \vspace{0.2cm}
        \caption{Prompt ICL Kind 9B}
        \label{fig:prompt_kind_icl_9b}
    \end{subfigure}%
    }
    \hfill
    \fbox{%
    \begin{subfigure}[t]{0.46\textwidth}
        \centering
        \includegraphics[width=0.95\linewidth]{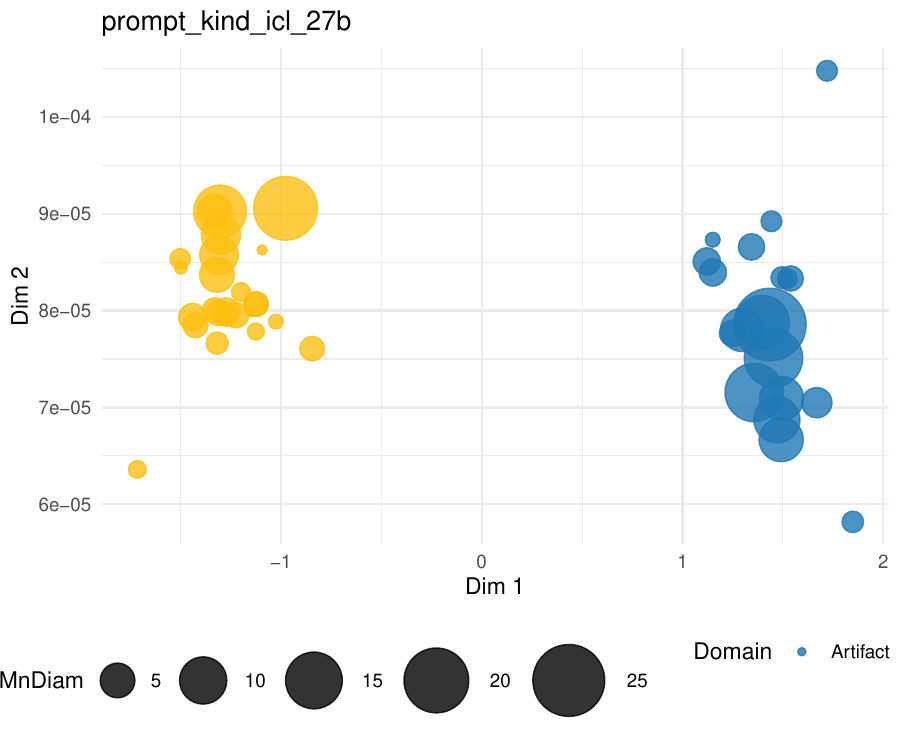}
        \vspace{0.2cm}
        \caption{Prompt ICL Kind 27B}
        \label{fig:prompt_kind_icl_27b}
    \end{subfigure}%
    }
    
    \vspace{0.8cm}
    
    \fbox{%
    \begin{subfigure}[t]{0.46\textwidth}
        \centering
        \includegraphics[width=0.95\linewidth]{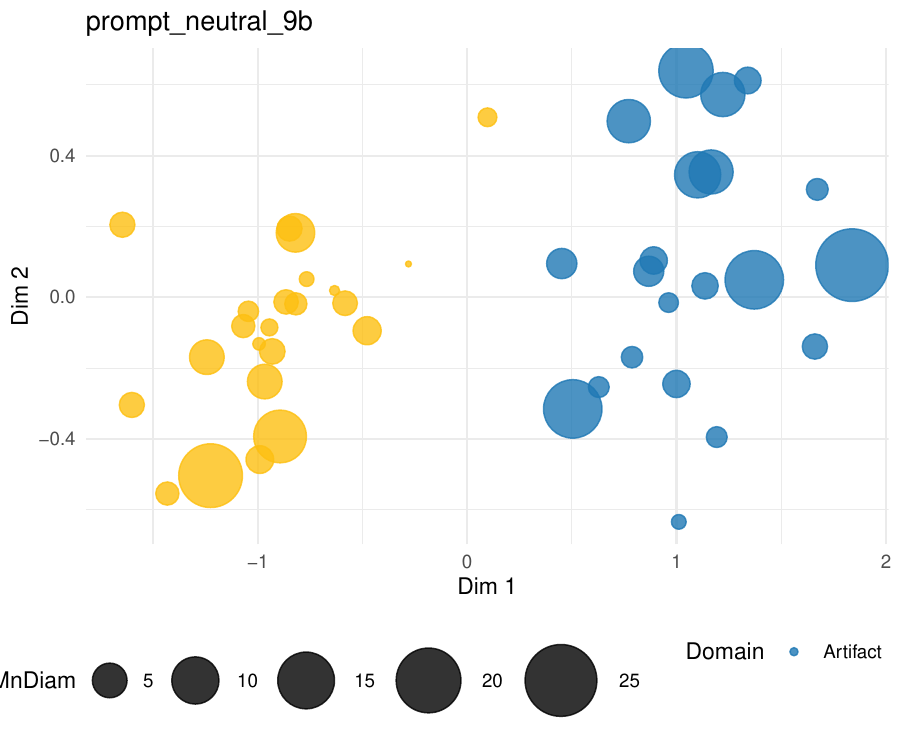}
        \vspace{0.2cm}
        \caption{Prompt Neutral 9B}
        \label{fig:prompt_neutral_9b}
    \end{subfigure}%
    }
    \hfill
    \fbox{%
    \begin{subfigure}[t]{0.46\textwidth}
        \centering
        \includegraphics[width=0.95\linewidth]{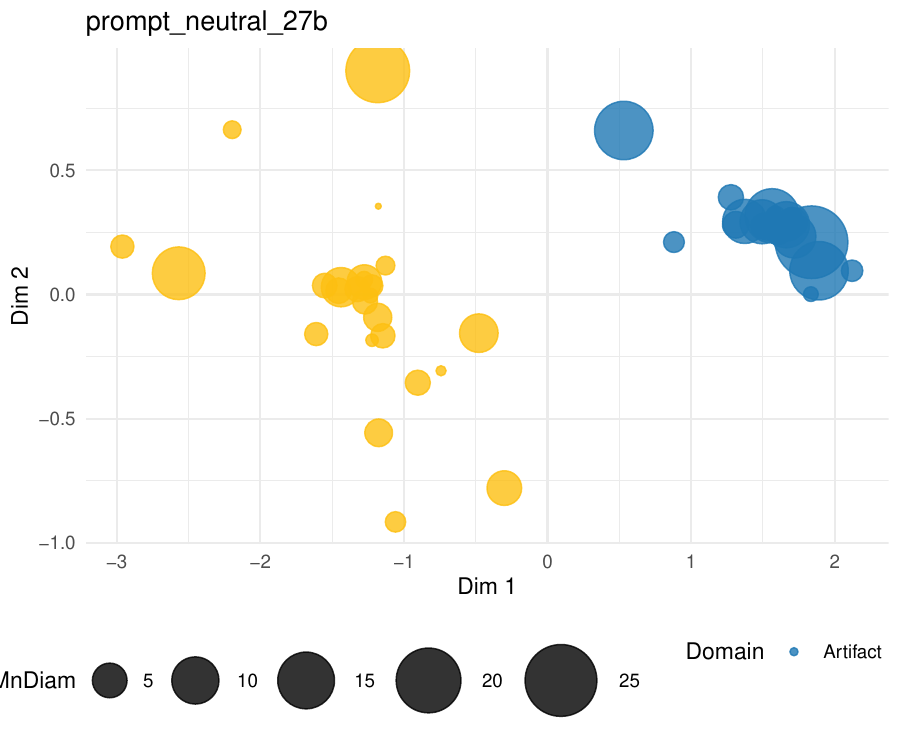}
        \vspace{0.2cm}
        \caption{Prompt Neutral 27B}
        \label{fig:prompt_neutral_27b}
    \end{subfigure}%
    }
    
    \vspace{0.8cm}
    
    \fbox{%
    \begin{subfigure}[t]{0.46\textwidth}
        \centering
        \includegraphics[width=0.95\linewidth]{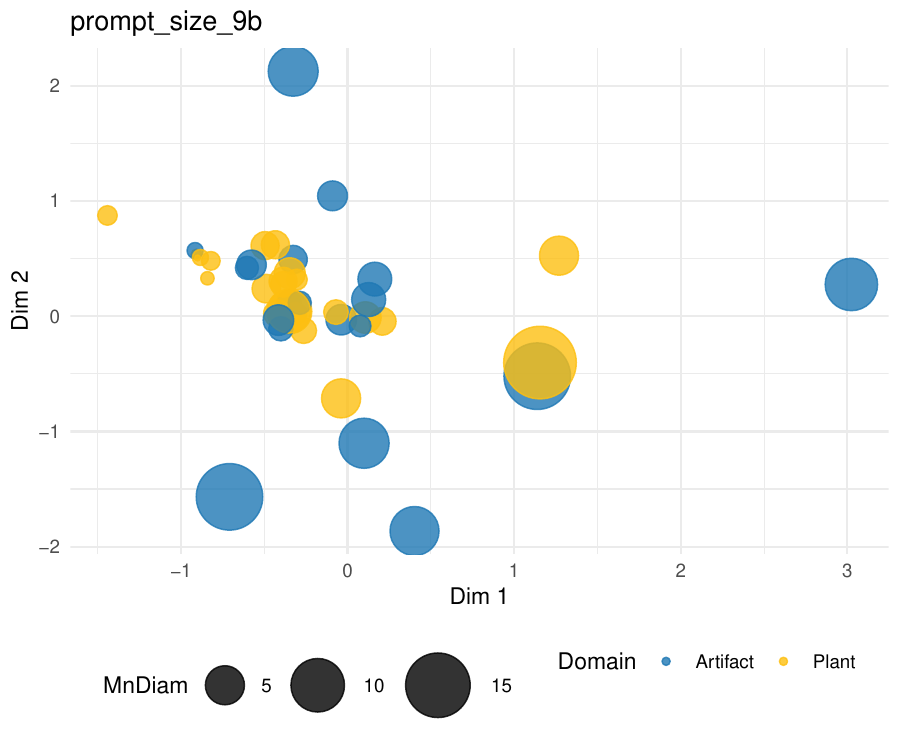}
        \vspace{0.2cm}
        \caption{Prompt Size 9B}
        \label{fig:prompt_size_9b}
    \end{subfigure}%
    }
    \hfill
    \fbox{%
    \begin{subfigure}[t]{0.46\textwidth}
        \centering
        \includegraphics[width=0.95\linewidth]{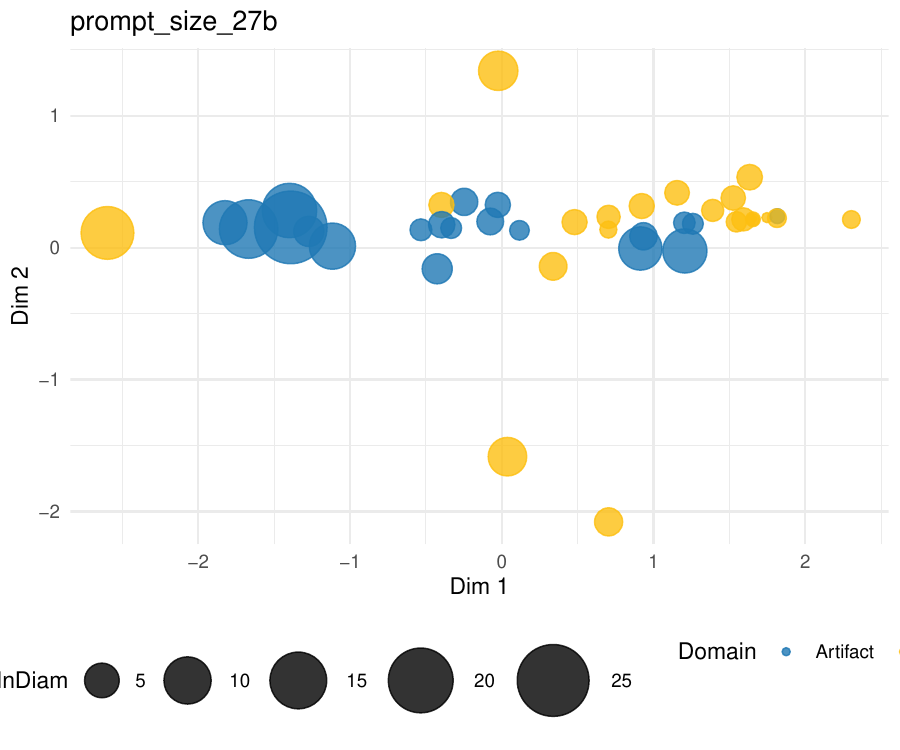}
        \vspace{0.2cm}
        \caption{Prompt Size 27B}
        \label{fig:prompt_size_27b}
    \end{subfigure}%
    }
    
    \vspace{0.5cm}
    \caption{Embedding Plots: Prompt Method Variations}
    \label{fig:embeddings_page2}
\end{figure}

\begin{figure}[p]
    \centering
    \vspace{0.5cm}
    
    \fbox{%
    \begin{subfigure}[t]{0.46\textwidth}
        \centering
        \includegraphics[width=0.95\linewidth]{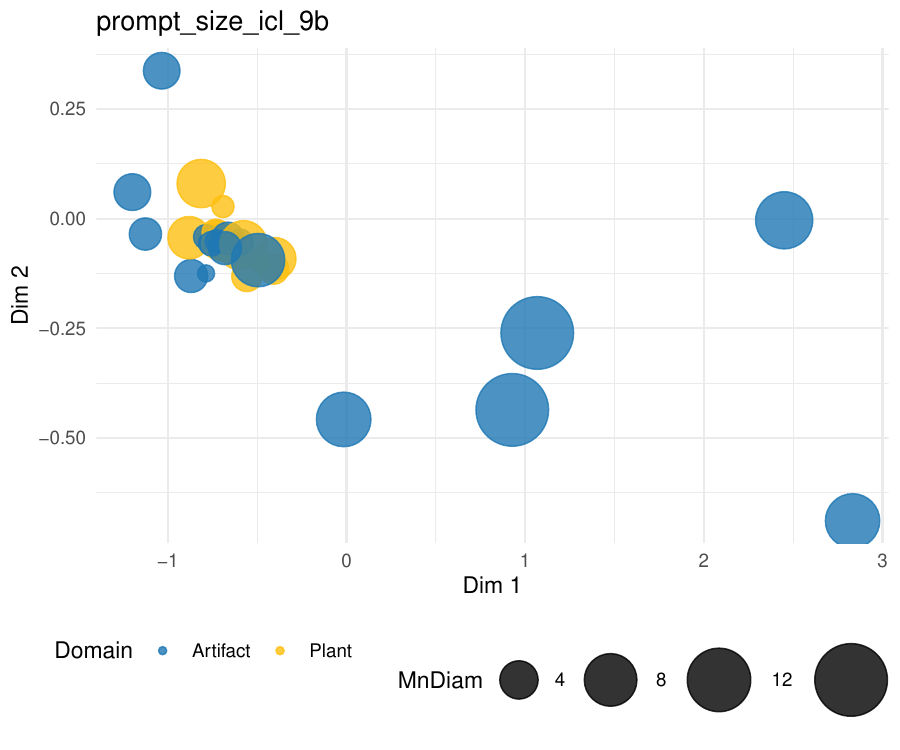}
        \vspace{0.2cm}
        \caption{Prompt ICL Size 9B}
        \label{fig:prompt_size_icl_9b}
    \end{subfigure}%
    }
    \hfill
    \fbox{%
    \begin{subfigure}[t]{0.46\textwidth}
        \centering
        \includegraphics[width=0.95\linewidth]{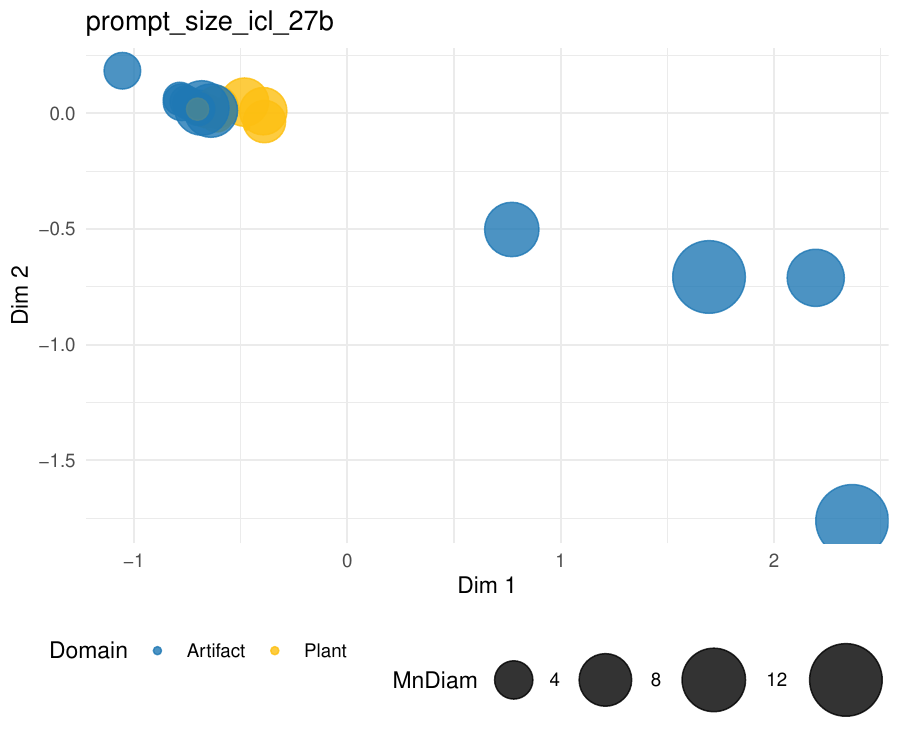}
        \vspace{0.2cm}
        \caption{Prompt ICL Size 27B}
        \label{fig:prompt_size_icl_27b}
    \end{subfigure}%
    }
    
    \vspace{0.8cm}
    
    \fbox{%
    \begin{subfigure}[t]{0.46\textwidth}
        \centering
        \includegraphics[width=0.95\linewidth]{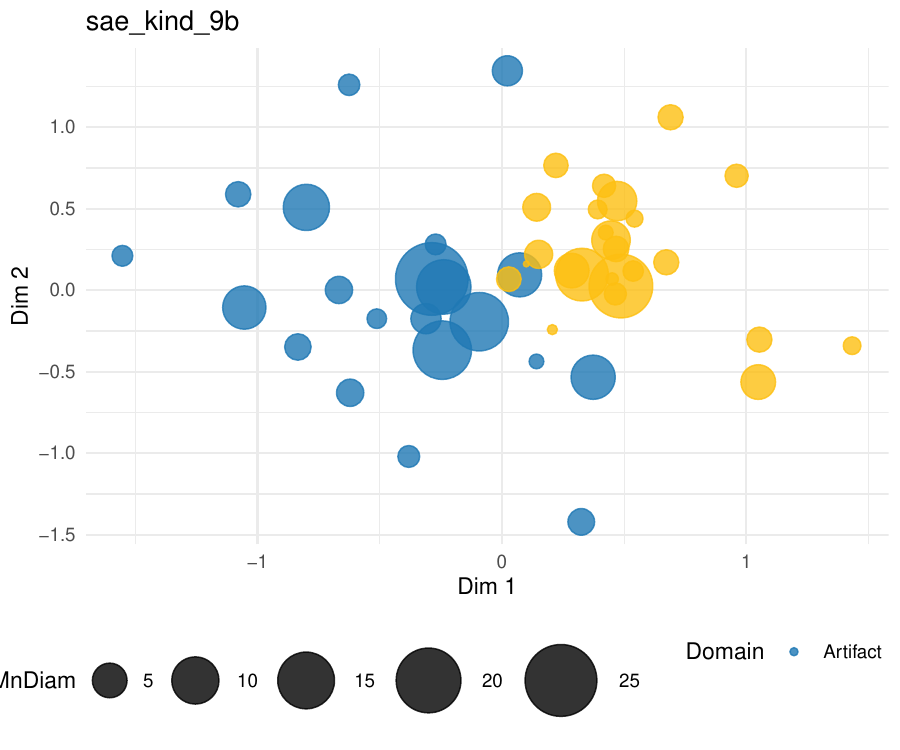}
        \vspace{0.2cm}
        \caption{SAE Kind 9B}
        \label{fig:sae_kind_9b}
    \end{subfigure}%
    }
    \hfill
    \fbox{%
    \begin{subfigure}[t]{0.46\textwidth}
        \centering
        \includegraphics[width=0.95\linewidth]{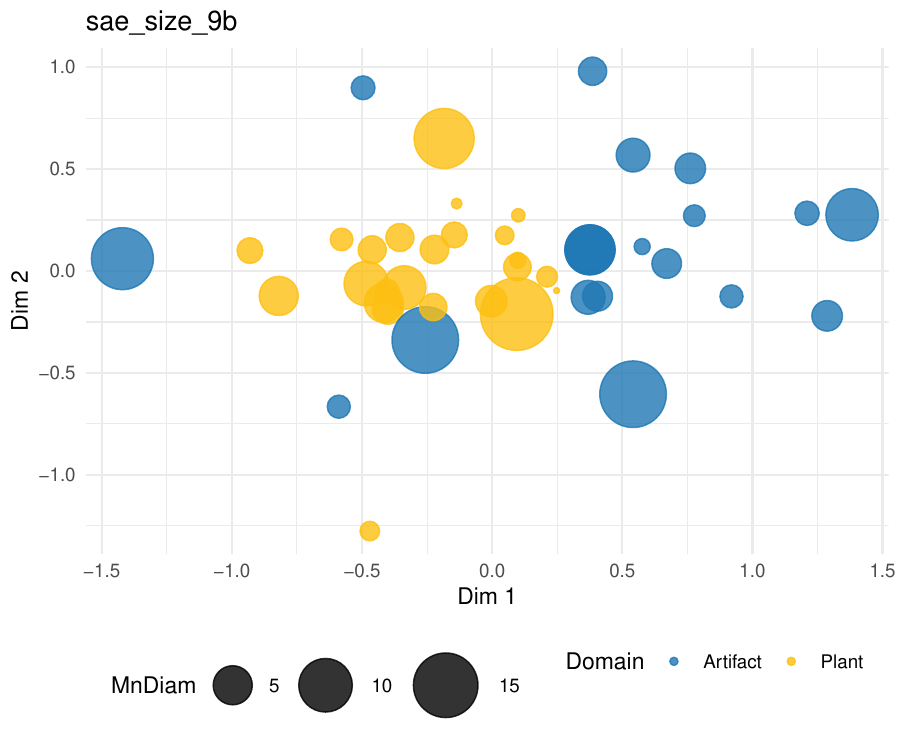}
        \vspace{0.2cm}
        \caption{SAE Size 9B}
        \label{fig:sae_size_9b}
    \end{subfigure}%
    }
    
    \vspace{0.8cm}
    
    \fbox{%
    \begin{subfigure}[t]{0.46\textwidth}
        \centering
        \includegraphics[width=0.95\linewidth]{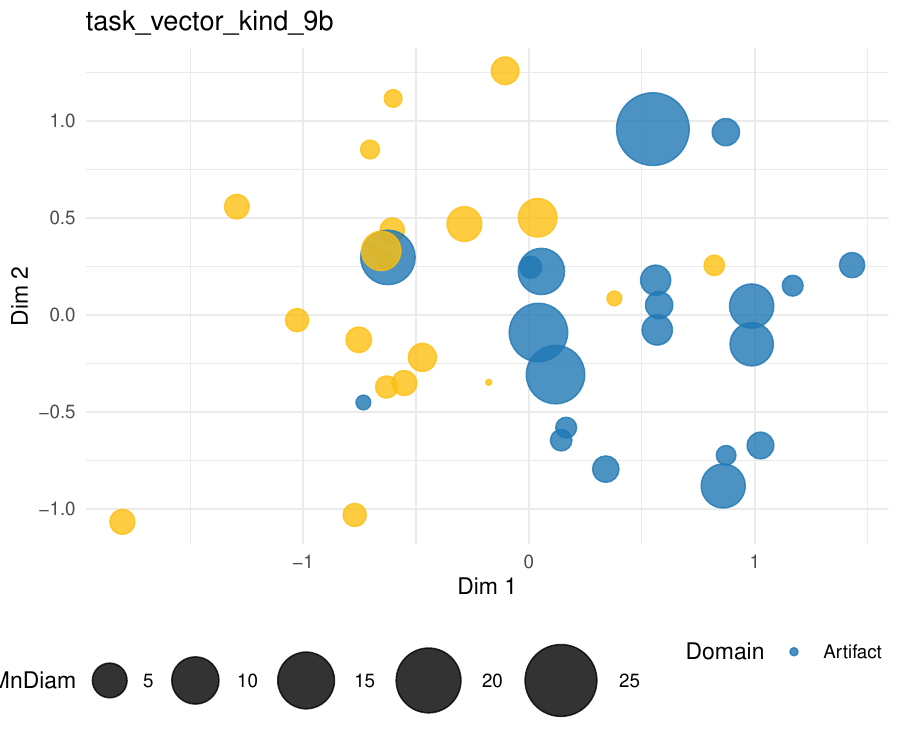}
        \vspace{0.2cm}
        \caption{Task Vector Kind 9B}
        \label{fig:task_vector_kind_9b}
    \end{subfigure}%
    }
    \hfill
    \fbox{%
    \begin{subfigure}[t]{0.46\textwidth}
        \centering
        \includegraphics[width=0.95\linewidth]{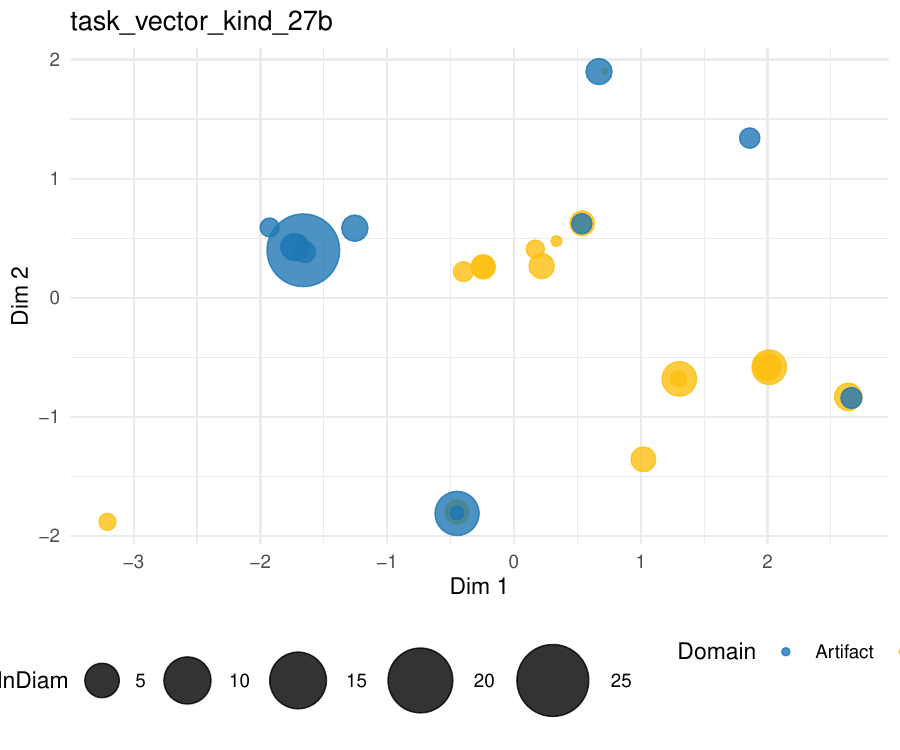}
        \vspace{0.2cm}
        \caption{Task Vector Kind 27B}
        \label{fig:task_vector_kind_27b}
    \end{subfigure}%
    }
    
    \vspace{0.5cm}
    \caption{Embedding Plots: Prompt ICL, SAE, and Task Vector Methods}
    \label{fig:embeddings_page3}
\end{figure}

\begin{figure}[p]
    \centering
    \vspace{0.5cm}
    
    \fbox{%
    \begin{subfigure}[t]{0.46\textwidth}
        \centering
        \includegraphics[width=0.95\linewidth]{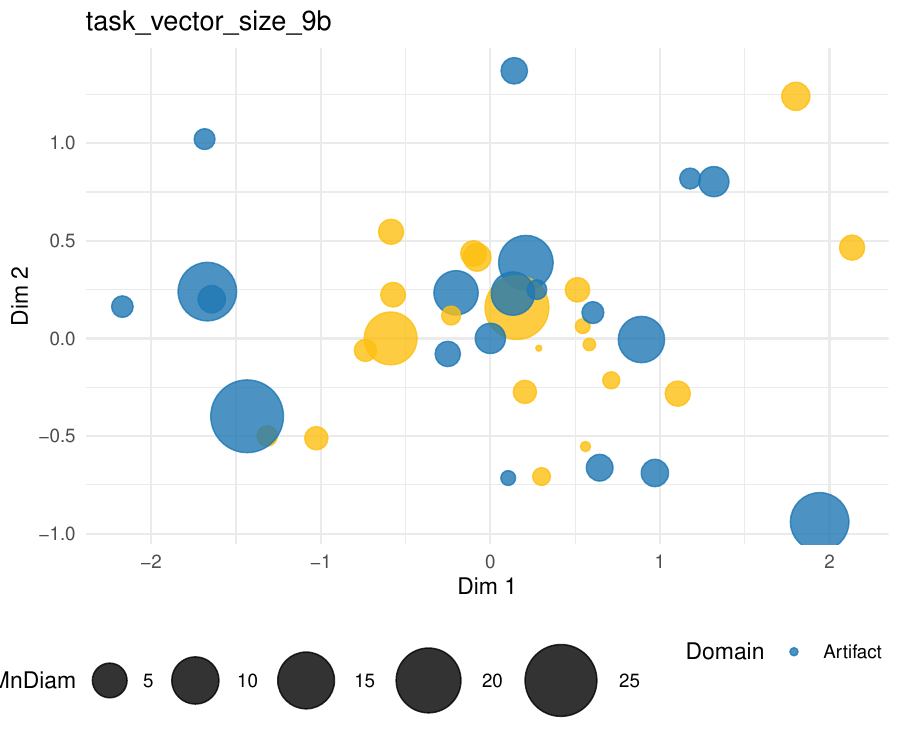}
        \vspace{0.2cm}
        \caption{Task Vector Size 9B}
        \label{fig:task_vector_size_9b}
    \end{subfigure}%
    }
    \hfill
    \fbox{%
    \begin{subfigure}[t]{0.46\textwidth}
        \centering
        \includegraphics[width=0.95\linewidth]{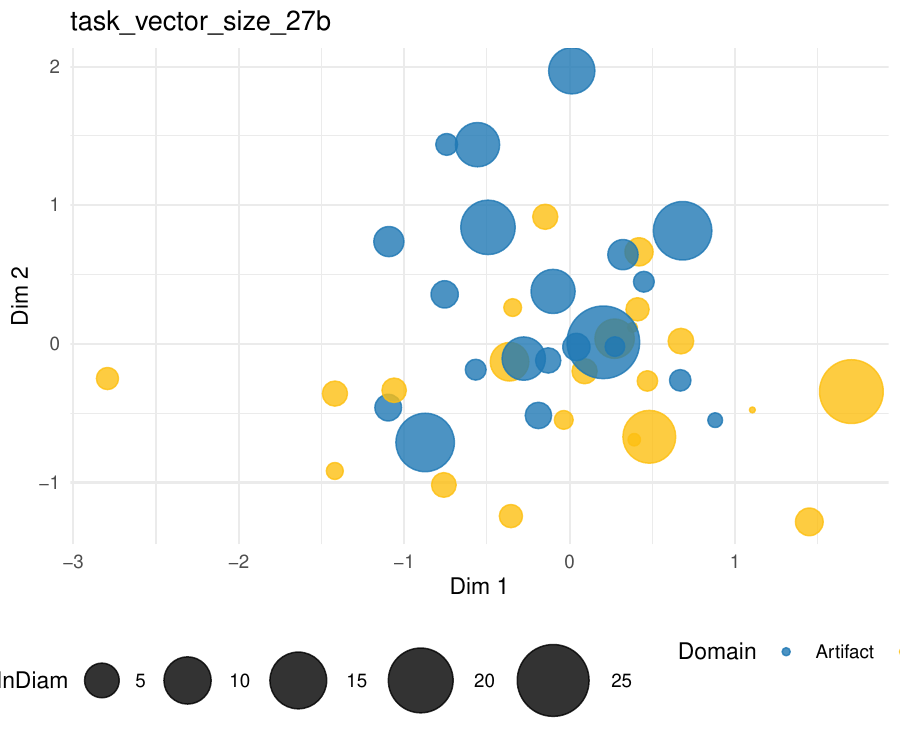}
        \vspace{0.2cm}
        \caption{Task Vector Size 27B}
        \label{fig:task_vector_size_27b}
    \end{subfigure}%
    }
    
    \vspace{2cm} 
    
    \caption{Embedding Plots: Task Vector Size Condition}
    \label{fig:embeddings_page4}
\end{figure}
\end{document}